\RequirePackage[svgnames, table]{xcolor}

\documentclass[11pt,letterpaper]{mystyle}
\usepackage[comma,authoryear,compress]{natbib}
\usepackage[all]{hypcap}
\hypersetup{colorlinks=true, citecolor=YaleBlue}

\usepackage{algorithm}
\usepackage{algorithmic}    

\usepackage{mathtools}
\usepackage{mathrsfs}
\usepackage{nicefrac}
\usepackage{dsfont}
\usepackage{bbm}
\usepackage{amsfonts}
\usepackage{fdsymbol}

\usepackage{subfig}       
\usepackage{wrapfig}
\usepackage{rotating}
\usepackage{multirow}
\usepackage{bigstrut}
\usepackage{makecell}
\usepackage{adjustbox}
\usepackage{stackengine}
\usepackage{float}
\usepackage{siunitx}

\usepackage{setspace}
\usepackage[normalem]{ulem}
\usepackage{afterpage}
\usepackage{lipsum}
\usepackage{pdfrender}

\usepackage{listings}
\tcbuselibrary{listings}

\usepackage[capitalize,noabbrev]{cleveref}

\definecolor{lightblue}{rgb}{0.22,0.45,0.70}
\definecolor{blanchedalmond}{rgb}{1.0, 0.92, 0.8}
\definecolor{carmine}{rgb}{0.59, 0.0, 0.09}
\definecolor{amaranth}{rgb}{0.9, 0.17, 0.31}
\definecolor{antiquebrass}{rgb}{0.8, 0.58, 0.46}
\definecolor{antiquefuchsia}{rgb}{0.57, 0.36, 0.51}
\definecolor{chromeyellow}{rgb}{0.31, 0.47, 0.26}
\definecolor{Gray}{gray}{0.95}
\definecolor{Cornsilk}{rgb}{1.0, 0.97, 0.86}
\definecolor{pink}{HTML}{fc6c85}
\definecolor{customblue}{HTML}{286dc0}
\definecolor{customred}{HTML}{CC0000}
\definecolor{customyellow}{HTML}{ffd55a}
\definecolor{customgrey}{HTML}{978d85}
\definecolor{diagblue}{RGB}{217,230,245}
\definecolor{diaggreen}{RGB}{216,237,204}

\tcbset{%
colback=gray!3,
colframe=black!60,
boxrule=0.6pt,
arc=2pt,
left=6pt,
right=6pt,
top=4pt,
bottom=4pt,
fonttitle=\bfseries
}

\tcbset{
aibox/.style={
  colback=gray!5,
  colframe=YaleBlue!70,
  boxrule=1pt,
  arc=3pt,
}
}
\newtcolorbox{AIbox}[2][]{aibox, title=#2, #1}

\newtcblisting{promptbox}[2][]{
  listing only,
  listing options={
      basicstyle=\ttfamily\footnotesize,
      breaklines=true,
  },
  colback=gray!10,
  colframe=YaleBlue!60,
  title=#2,
  fonttitle=\bfseries,
  #1
}

\newtcolorbox{blueBox}[1][]{
colback=YaleBlue!12!white,
colframe=YaleBlue!85,
boxrule=1pt,
arc=3pt,
floatplacement=floating,
title=\centering #1
}

\newtcolorbox{yellowBox}[1][]{
colback=customyellow!10!white,
colframe=customyellow,
floatplacement=floating,
title=\centering #1
}

\newtcolorbox{promptBox}[1][]{
colback=YaleBlue!12!white,
colframe=YaleBlue!85,
boxrule=0.8pt,
arc=3pt,
floatplacement=floating,
title=\centering #1
}

\newtcolorbox{wronganswer}[1][]{
  enhanced,
  breakable,
  colframe=customred!65,
  colback=customred!6!white,
  sharp corners,
  boxsep=0pt,
  left=5pt,
  right=5pt,
  top=6pt,
  bottom=6pt,
  boxrule=0pt,
  leftrule=4pt,
  #1
}

\newtcolorbox{correctanswer}[1][]{
  enhanced,
  breakable,
  colframe=OliveGreen,
  colback=OliveGreen!10!white,
  sharp corners,
  boxsep=0pt,
  left=5pt,
  right=5pt,
  top=6pt,
  bottom=6pt,
  boxrule=0pt,
  leftrule=4pt,
  #1
}

\setlist[itemize]{leftmargin=*}
\setlist[enumerate]{leftmargin=*}

\newcolumntype{a}{>{\columncolor{Gray}}c}

\renewcommand{\mathbf}{\boldsymbol}

\newcommand{\w}{\mathbf{w}}

\makeatletter
\def\Ddots{\mathinner{\mkern1mu\raise\p@
\vbox{\kern7\p@\hbox{.}}\mkern2mu
\raise4\p@\hbox{.}\mkern2mu\raise7\p@\hbox{.}\mkern1mu}}

\makeatother

\title{Context Bootstrapped Reinforcement Learning}
\runningtitle{Context Bootstrapped Reinforcement Learning}

\newcommand{\cisco}{2}
\newcommand{\ucsb}{1}

\author[\ucsb]{Saaket Agashe}
\author[\cisco]{Jayanth Srinivasa}
\author[\cisco]{Gaowen Liu}
\author[\cisco]{Ramana Kompella}
\author[\ucsb]{Xin Eric Wang}

\affil[\ucsb]{University of California, Santa Barbara}
\affil[\cisco]{Cisco Research}

\begin{document} 

\begin{abstract}
Reinforcement Learning from Verifiable Rewards (RLVR) suffers from exploration inefficiency, where models struggle to generate successful rollouts, resulting in minimal learning signal. This challenge is particularly severe for tasks that require the acquisition of novel reasoning patterns or domain-specific knowledge. To address this, we propose \textbf{Context Bootstrapped Reinforcement Learning (CBRL)}, which augments RLVR training by stochastically prepending few-shot demonstrations to training prompts. The injection probability follows a curriculum that starts high to bootstrap early exploration, then anneals to zero so the model must ultimately succeed without assistance. This forces the policy to internalize reasoning patterns from the demonstrations rather than relying on them at test time. We validate CBRL across two model families and five Reasoning Gym tasks. Our results demonstrate that CBRL consistently improves success rate, provides better exploration efficiency, and is algorithm-agnostic. We further demonstrate CBRL's practical applicability on Q, a domain-specific programming language that diverges significantly from mainstream language conventions.
\vspace{3mm}

\parbox{\linewidth}{
\textbf{Correspondence:} saaket@ucsb.edu, ericxwang@ucsb.edu\\ 
\textbf{Project Website:} \url{https://context-bootstrapped-rl.github.io}
}
\end{abstract}

\maketitle

\section{Introduction}

Reinforcement Learning from Verifiable Rewards (RLVR)~\citep{tulu3} has emerged as a powerful paradigm for post-training Large Language Models (LLMs) on challenging reasoning problems~\citep{deepseekr1}. In RLVR, the model receives a binary reward purely based on its final output. This learning framework has driven remarkable advances across diverse domains, including mathematical reasoning~\citep{deepseekr1, yang2024qwen2}, code generation~\citep{jiang2025coderl+}, and tool use~\citep{feng2025retool}.

However, RLVR in Large Language Models suffers from a fundamental problem: Exploration Inefficiency~\citep{liu2025evocotovercomingexplorationbottleneck, yue2025limit, drgrpo}. When a model cannot reliably produce correct rollouts during training, it receives minimal learning signal, leading to slow or failed convergence. This challenge is particularly acute in settings where the domain is underrepresented in pretraining data (e.g., domain-specific programming languages) or because the task requires novel reasoning patterns.

Therefore, we propose \emph{Context Bootstrapped Reinforcement Learning (CBRL)}, a novel approach that leverages the In-Context Learning~\citep{fewshot} capabilities of LLMs to address exploration inefficiency in RLVR. The method maintains a small bank of few-shot examples that are dynamically injected into training prompts with decreasing probability as training progresses. Early in training, In-Context examples guide the model toward successful rollouts, providing the learning signal needed to bootstrap capability acquisition. As training advances, the injection probability anneals, encouraging the model to perform independently.

We validate CBRL through extensive experiments on two model families (Qwen~\citep{qwen} and Llama~\citep{llama}) across five Reasoning Gym tasks~\citep{DBLP:journals/corr/abs-2505-24760}: ARC-1D, Word Sorting, Spell Backward, Matrix Manipulation, and Puzzle-24. These
tasks require the model to learn specific reasoning patterns. We further evaluate on Q Programming~\citep{DBLP:journals/corr/abs-2508-06813}, a domain-specific language for time-series databases. Unlike mainstream languages, Q features right-to-left evaluation,  implicit typing, and a terse array-oriented syntax that diverges from conventions dominant in pretraining corpora. Our results demonstrate that CBRL consistently outperforms the GRPO-only baseline, improving accuracy on all 10 model–environment pairs (Table~\ref{tab:sr_results_both_models_mean_se}), with gains ranging from +1.3\% to +22.3\%. On Q programming, CBRL improves both test-pass rate ($27.3\% \rightarrow 43.0\%$) and Pass@1 ($5.0\% \rightarrow 26.3\%$) over the regular GRPO (Table~\ref{tab:q_lang_results}).

We find that CBRL is \emph{algorithm-agnostic}, yielding consistent gains when paired with both GRPO~\citep{deepseekmath} and RLOO~\citep{DBLP:conf/acl/AhmadianCGFKPUH24}. Under RLOO, Word Sorting improves from 20\% to 67\% and Puzzle-24 from 23\% to 66\% (Table~\ref{tab:rloo_mean_se_qwen_simple}). Training curve analysis reveals that CBRL addresses the \emph{exploration inefficiency} bottleneck, attaining substantially higher mean reward early in training, consistent with improved exploration from injected few-shot demonstrations (Figure~\ref{fig:training_curves}). Crucially, this advantage persists as the injection probability $p_i$ anneals toward zero. Performance does not collapse once demonstrations are removed, indicating that CBRL bootstraps durable behaviors rather than inducing dependence on in-context examples. We also ablate the initial injection probability and find that moderate values around $p_i = 0.5$ perform best. Setting $p_i$ too high overwhelms the policy's own exploration, while setting it too low provides insufficient scaffolding to escape reward plateaus.

\textbf{Our main contributions are:}
\begin{itemize}
\item We introduce Context Bootstrapped Reinforcement Learning (CBRL), an algorithm-agnostic approach that leverages in-context learning to address exploration inefficiency in RLVR. 
\item We validate CBRL across five Reasoning Gym tasks and the domain-specific Q programming language, demonstrating consistent gains across diverse tasks and domains.
\item We provide an in-depth analysis of CBRL, including learning dynamics (training curves), ablations over the injection-probability schedule, and qualitative examples that characterize its usefulness. 
\end{itemize}


\section{Context Bootstrapped Reinforcement Learning}

\begin{figure}[t]
    \centering
    \subfloat[]{
        \includegraphics[width=0.52\textwidth]{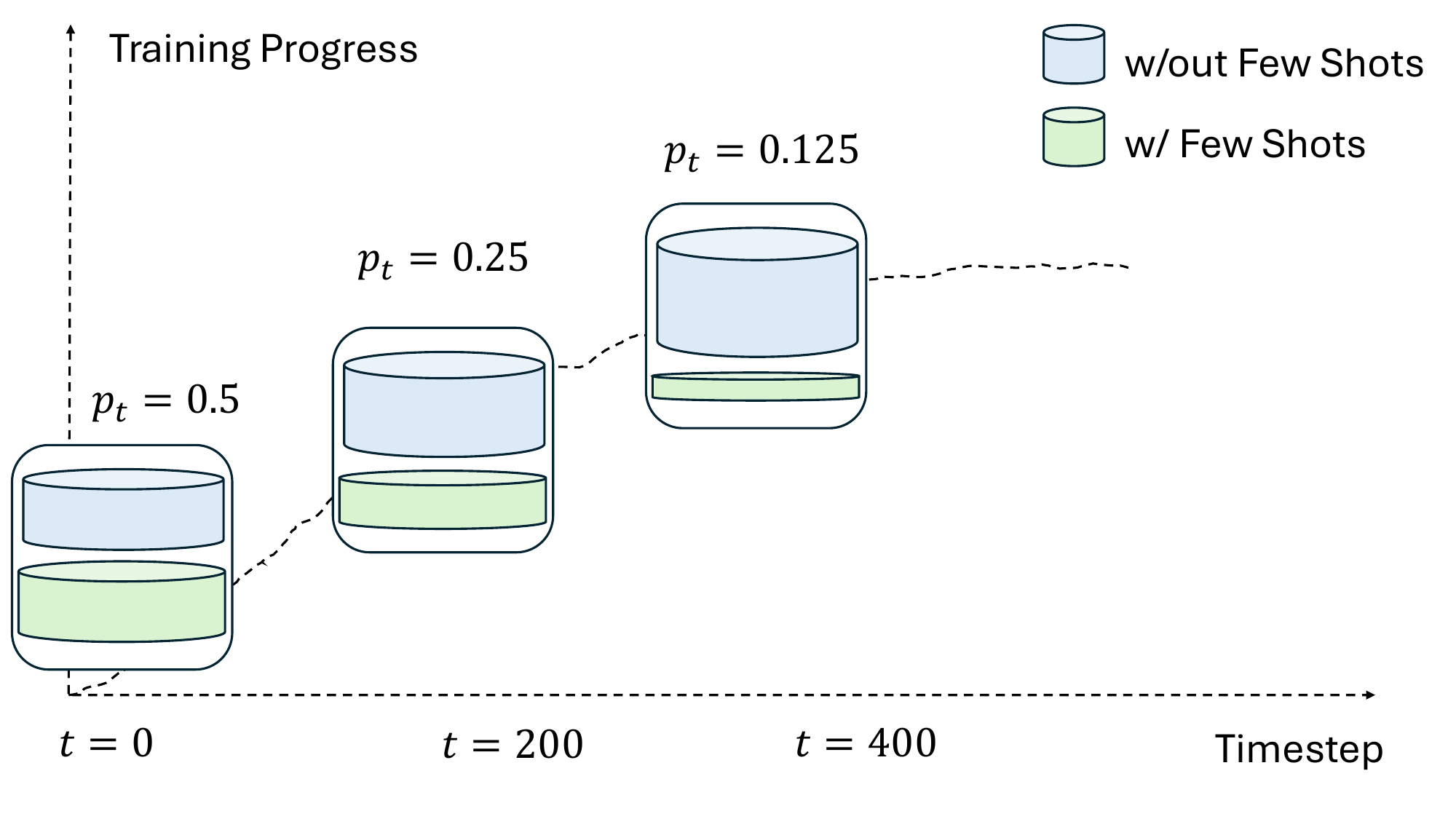}
    }\hfill
    \subfloat[]{
        \includegraphics[width=0.45\textwidth]{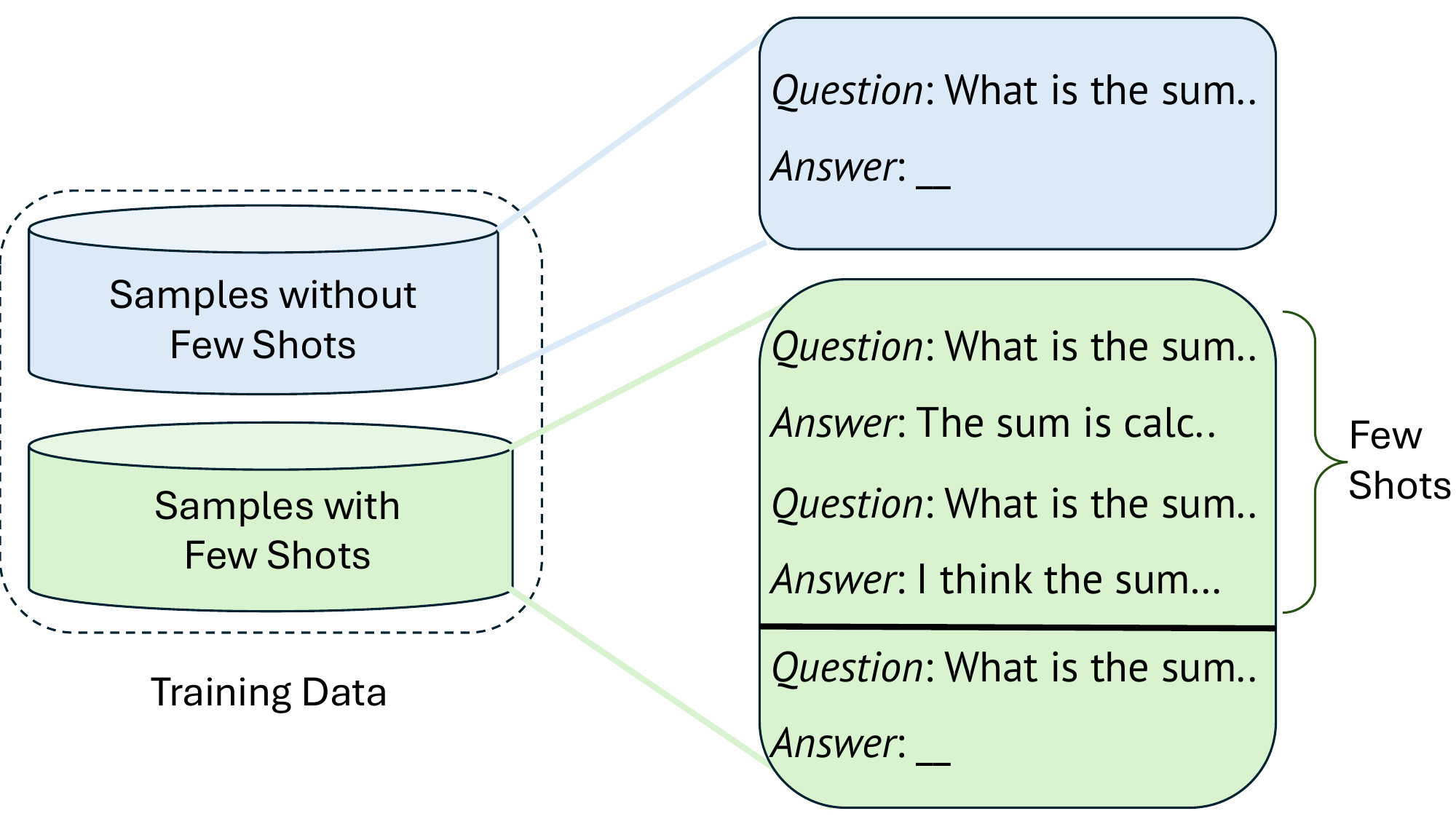}
    }
    \caption{Overview of the Context Bootstrapped Reinforcement Learning (CBRL) framework. \textbf{(a)}~The injection probability annealing schedule. At each timestep~$t$, training batches consist of {\setlength{\fboxsep}{1pt}\colorbox{diagblue}{samples without few shots}} and {\setlength{\fboxsep}{1pt}\colorbox{diaggreen}{samples with few shots}}, with the proportion governed by~$p_i$, which decreases linearly from~$p_{\mathrm{start}}$ to~$p_{\mathrm{end}}$. \textbf{(b)}~Construction of the two sample types. {\setlength{\fboxsep}{1pt}\colorbox{diagblue}{Samples without exemplars}} (top) present only the task query followed by an empty assistant turn. {\setlength{\fboxsep}{1pt}\colorbox{diaggreen}{Samples with exemplars}} (bottom) prepend solved few-shot demonstrations as prior user--assistant exchanges before the target query.
    }
    \label{fig:cbrl-overview}
\end{figure}

Context Bootstrapped Reinforcement Learning (CBRL) augments the standard RLVR training process by incorporating in-context examples as temporary scaffolding. The method operates through three core components: a bank of solved examples, a stochastic injection mechanism, and a curriculum schedule that phases out assistance as training progresses. At inference time, the trained policy generates completions without any injected examples, incurring no additional computational cost. Figure~\ref{fig:cbrl-overview} provides an overview of the framework.

\subsection{Few-Shot Example Bank}

For each task domain, we construct a bank $\mathcal{B}$ of solved examples, where each $e \in \mathcal{B}$ contains a problem statement $q$, an optional reasoning trace $r$, and an answer $a$. The bank can be constructed from expert demonstrations, stronger model solutions, or hand-crafted instances.

\subsection{Stochastic Context Injection}

At each training step, with probability $p_i$, we sample $k$ examples $\{e_1, \ldots, e_k\}$ from the bank $\mathcal{B}$ and prepend them to the query as conversation pairs. The policy generates completions on this augmented prompt, with rewards computed solely on the generated response. As illustrated in Figure~\ref{fig:cbrl-overview}b, samples without exemplars present only the task query followed by an empty assistant turn, while samples with exemplars prepend solved few-shot demonstrations as prior user-assistant exchanges before the target query.

This stochastic injection ensures the policy experiences both standalone and ICL-augmented prompts. The mixture serves a critical purpose. If examples appeared on every prompt, the policy might learn to depend on their presence rather than internalize the demonstrated reasoning patterns. By randomly withholding examples, we force the model to attempt problems independently, even during early training.

\subsection{Curriculum Schedule}

The injection probability $p_i$ follows a linear annealing schedule:
\begin{equation}
p_i = p_{\text{start}} + \frac{t-1}{T-1}(p_{\text{end}} - p_{\text{start}})
\end{equation}
where $p_{\text{start}}$ is the initial injection probability (typically 0.5 to 1.0), $p_{\text{end}}$ is the final injection probability (typically 0.0), $T$ is the total training budget, and $t$ is the current training step.

This schedule addresses a fundamental tension in learning from demonstrations. Early in training, when the policy is weakest and would otherwise fail to discover correct solutions, examples appear frequently ($p \approx p_{\text{start}}$), providing strong guidance toward effective solution strategies. However, persistent scaffolding would prevent the model from developing independent problem-solving capabilities. As training progresses and the policy improves, injection frequency decreases, requiring increasingly autonomous performance. By the end of training ($t = T$), injection probability reaches $p_{\text{end}} = 0$, and the policy must rely entirely on its internalized knowledge. Figure~\ref{fig:cbrl-overview}a visualizes this annealing process, showing how the proportion of exemplar-augmented samples in each training batch decreases over time. The annealing rate automatically adjusts to any training budget $T$, making the curriculum adaptive across different experimental configurations.

Algorithm~\ref{alg:cbrl} presents the complete CBRL training pipeline.

\begin{algorithm}[h]
\caption{Context Bootstrapped Reinforcement Learning}
\label{alg:cbrl}
\begin{algorithmic}[1]
\STATE \textbf{Input:} base policy $\pi_\theta$, training tasks $\mathcal{T}$, few-shot bank $\mathcal{B}$, annealing schedule $\{p_i\}_{t=1}^T$, bank sample size $k$, mini-batch size $m$
\vspace{3pt}\hrule\vspace{3pt}
\FOR{$t = 1$ to $T$}
    \STATE Draw mini-batch $\{q_i\}_{i=1}^m \sim \mathcal{T}$
    \STATE Set injection probability $p \leftarrow p_i$
    \vspace{2pt}
    \STATE \textit{// Context injection}
    \FOR{each $q_i$}
        \STATE $E_i \leftarrow \textsc{Sample}(\mathcal{B}, k)$
        \STATE $b_i \leftarrow \text{Bernoulli}(p)$
        \STATE $x_i \leftarrow \textsc{Compose}(b_i, E_i, q_i)$
    \ENDFOR
    \vspace{2pt}
    \STATE \textit{// Policy optimization}
    \STATE $\mathcal{D}_t \leftarrow \textsc{RolloutBatch}(\pi_\theta, \{x_i\}_{i=1}^m)$
    \STATE $\pi_\theta \leftarrow \textsc{PolicyUpdate}(\pi_\theta, \mathcal{D}_t)$
\ENDFOR
\vspace{3pt}\hrule\vspace{3pt}
\STATE \textbf{return} $\pi_\theta$
\end{algorithmic}
\end{algorithm}

\subsection{Compatibility with Policy Gradient Methods}

A key advantage of CBRL is its algorithm-agnostic design. The method modifies only the input distribution during training, augmenting prompts with in-context examples, without altering the underlying RL objective, loss functions, or optimization procedure. From the optimizer's perspective, CBRL simply changes the input space, treating augmented prompts as another form of training data. This transparency makes CBRL trivially compatible with any policy gradient method (GRPO, RLOO, or others) and allows it to compose with other training enhancements.


\section{Experimental Setup}

We evaluate Context Bootstrapped Reinforcement Learning across two experimental settings: five reasoning tasks from the Reasoning Gym benchmark, and code generation in the Q programming language. These settings test CBRL's effectiveness on both synthetic reasoning challenges as well as a real-world domain with unconventional syntax that is unfamiliar to LLMs.

\subsection{Environments}

\textbf{Reasoning Gym.} We train and evaluate on five reasoning environments from Reasoning Gym~\citep{DBLP:journals/corr/abs-2505-24760}. Each environment generates problems procedurally, providing unlimited training data. The generations can be controlled using seeds.

\begin{itemize}
    \item \textit{ARC-1D}: A one-dimensional adaptation of the ARC-AGI challenge~\citep{DBLP:journals/corr/abs-2505-11831}. Each problem presents $k \in \{2, 3\}$ input-output examples of integer arrays (values in $[0, 9]$, lengths 10--30) and requires inferring the transformation rule to predict the output for a held-out input.
    \item \textit{Manipulate Matrix}: Problems specify a sequence of matrix operations (transpose, rotate, flip, slice, etc.) to apply to an initial integer grid, with grids ranging from $2 \times 2$ to $10 \times 10$ and 1--10 operations per problem.
    \item \textit{Word Sorting}: Problems provide a list of 3--10 randomly generated words (3--12 characters each) that must be sorted alphabetically.
    \item \textit{Spell Backward}: Problems provide a single randomly generated word (3--10 characters) that must be reversed character-by-character.
    \item \textit{Puzzle-24}: The classic ``24 game'' where four integers (each in $[1, 9]$) must be combined using $+, -, \times, \div$ to produce exactly 24. All generated problems are verified to have valid solutions.
\end{itemize}

\begin{figure*}[t]
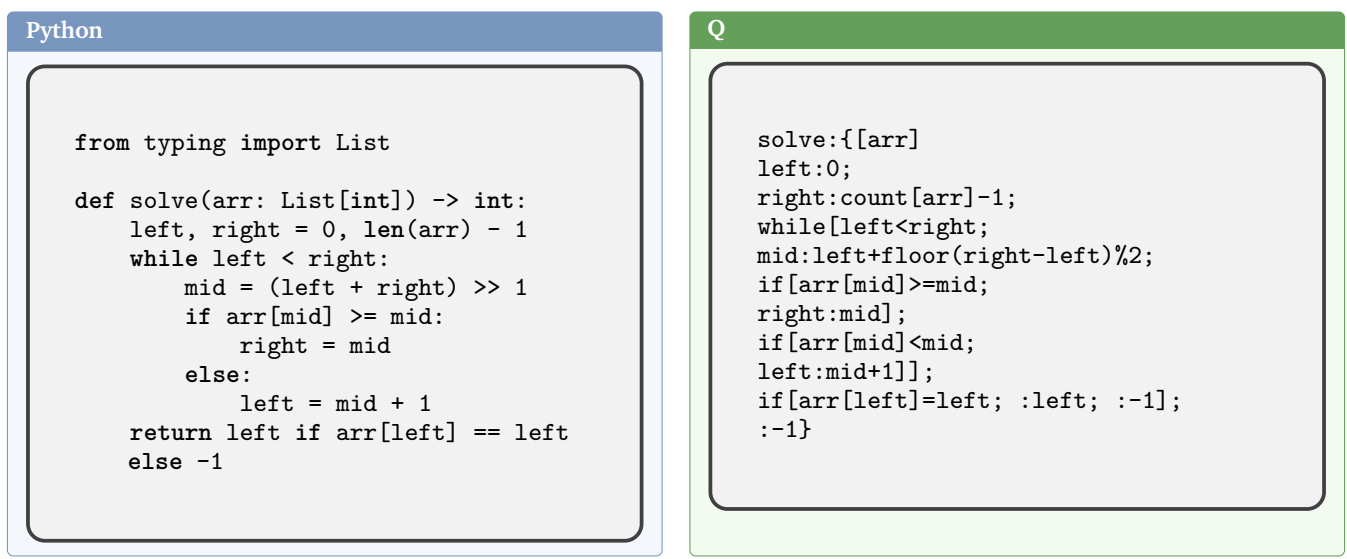


\definecolor{boxblue}{RGB}{217,230,245}
\definecolor{boxgreen}{RGB}{216,237,204}
\definecolor{borderblue}{RGB}{120,150,190}
\definecolor{bordergreen}{RGB}{100,160,90}

    \lstdefinestyle{code}{
      basicstyle=\ttfamily\footnotesize,
      columns=fullflexible,
      keepspaces=true,
      showstringspaces=false,
      breaklines=true,
      tabsize=2,
    }
    
    \tcbset{
      qualstyle/.style={
        enhanced,
        fonttitle=\bfseries\footnotesize,
        fontupper=\footnotesize,
        boxrule=0.4pt,
        arc=2pt,
        top=2pt, bottom=2pt, left=4pt, right=4pt,
        before upper={\setlength{\parskip}{2pt}},
      }
    }

  \centering

  \begin{minipage}[t]{0.49\linewidth}\vspace{0pt}
  \begin{tcolorbox}[qualstyle,
    colback=boxblue!30, colframe=borderblue,
    colbacktitle=borderblue, coltitle=white,
    title=Python,
    equal height group=twoblocks]
\begin{tcblisting}{listing only, listing options={style=code, language=Python}}
from typing import List

def solve(arr: List[int]) -> int:
    left, right = 0, len(arr) - 1
    while left < right:
        mid = (left + right) >> 1
        if arr[mid] >= mid:
            right = mid
        else:
            left = mid + 1
    return left if arr[left] == left else -1
\end{tcblisting}
  \end{tcolorbox}
  \end{minipage}\hfill
  \begin{minipage}[t]{0.49\linewidth}\vspace{0pt}
  \begin{tcolorbox}[qualstyle,
    colback=boxgreen!30, colframe=bordergreen,
    colbacktitle=bordergreen, coltitle=white,
    title=Q,
    equal height group=twoblocks]
\begin{tcblisting}{listing only, listing options={style=code}}
solve:{[arr]
left:0;
right:count[arr]-1;
while[left<right;
mid:left+floor(right-left)%2;
if[arr[mid]>=mid;
right:mid];
if[arr[mid]<mid;
left:mid+1]];
if[arr[left]=left; :left; :-1];
:-1}
\end{tcblisting}
  \end{tcolorbox}
  \end{minipage}

  \caption{Equivalent implementations of a fixed-point search algorithm in Python (left) and Q (right). Both solutions use binary search to find an index $i$ such that $\texttt{arr}[i] = i$. While the Python version follows conventional syntax familiar to pretrained language models, the Q implementation employs a terse, array-oriented style with right-to-left evaluation, minimal punctuation, and implicit returns.}
  \label{fig:q_example}
\end{figure*}

\textbf{Q Programming.} Q is a vector-based programming language for financial applications built on the kdb+ database system. Its specialized syntax and limited representation in public codebases make it a natural testbed for exploration inefficiency evaluations, as LLMs have limited prior exposure to Q code. We use the dataset and evaluation framework from \citet{DBLP:journals/corr/abs-2508-06813}, which contains 678 Leetcode-style Q programming problems (542 train and 136 test) spanning topics such as arrays, search, sorting, and dynamic programming. Figure~\ref{fig:q_example} shows a side-by-side comparison of a fixed-point search function implemented in Python and Q.

\subsection{Models}

For Reasoning Gym, we use Qwen2.5-3B-Instruct~\citep{qwen} and Llama-3.2-3B-Instruct~\citep{llama} as base models. Qwen2.5-3B-Instruct provides a controlled evaluation setting: it achieves a near-zero baseline. We additionally evaluate on Llama-3.2-3B-Instruct to test generalization across model families and to control for specific training dynamics observed in Qwen models~\citep{DBLP:journals/corr/abs-2506-10947}. 

For Q programming, we use the QQwen-7B-Pretrain~\citep{DBLP:journals/corr/abs-2508-06813} model, a Qwen-based model pretrained on Q language corpora. This provides a stronger starting point than general-purpose LLMs while still exhibiting limited Q programming ability before RLVR.

\subsection{Training}

\paragraph{Group Relative Policy Optimization (GRPO).} We train using Group Relative Policy Optimization~\citep{deepseekmath, deepseekr1}. For Reasoning Gym, we train for 500 steps with a batch size of 32 using FSDP, providing a +0.2 reward for correct formatting and
  +1.0 for the correct final answer. For Q programming, we use a batch size of 64 and group size of 8, with partial rewards proportional to test case accuracy (e.g., +0.4 for 2/5 correct) and a +2 bonus when all test cases pass.

  \paragraph{Context Bootstrapped Reinforcement Learning.} We augment GRPO with context bootstrapping by stochastically prepending few-shot demonstrations during training. The injection probability starts at $p_0 = 0.5$ and decays linearly to $0$ over the course of
  training. When injection occurs, we prepend $k=2$ examples from a task-specific few-shot bank. The bank construction and selection strategy differ by task, as described below.

  \paragraph{Few-Shot Banks.} For Reasoning Gym, we construct a bank of 20 problems per task, solved programmatically to obtain ground-truth answers. GPT-5.2 generates step-by-step reasoning traces for each solution, yielding entries of the form $\langle q, r, a
  \rangle$ (question, reasoning, answer). At injection time, examples are sampled uniformly at random.

  For Q programming, the bank consists of 50 verified code examples drawn from the SFT dataset, containing only code without reasoning annotations. Rather than random sampling, we filter candidates at injection time to include only examples sharing tags (e.g.,
  \texttt{Array}, \texttt{Dynamic Programming}) with the current training problem, then sample from this filtered set.

\subsection{Evaluation}

For Reasoning Gym, we evaluate using the programmatic verifier, which returns 1.0 for exact matches and 0.0 otherwise. We test on 100 held-out problems per environment, generated with different random seeds than the training samples. Reported results are the average of three runs. 

For Q programming, correctness is measured using five unit tests per problem, evaluated on the standard test split from \citet{DBLP:journals/corr/abs-2508-06813}. Reported results are the average of five runs.

\section{Results and Analysis}
\subsection{Main Results}
\begin{table*}[t]
  \centering
  \caption{Success rates (\%) on five Reasoning Gym environments for Qwen2.5-3B-Instruct and Llama-3.2-3B-Instruct. \textbf{Baseline} and \textbf{Baseline w/ Few Shots} evaluate the pretrained model without any RL training (zero-shot and few-shot prompting, respectively). \textbf{GRPO} is standard GRPO training; \textbf{CBRL GRPO} augments GRPO with context bootstrapping (ours). All values are mean $\pm$ standard error over three seeds; bold indicates the best result per environment and model. CBRL improves over standard GRPO in all ten model--environment pairs, with gains ranging from $+1.3\%$ (Spell Backward, Llama) to $+22.3\%$ (Word Sorting, Qwen).}
  \label{tab:sr_results_both_models_mean_se}
  \renewcommand{\arraystretch}{1.25}
  \resizebox{\textwidth}{!}{%
  \begin{tabular}{@{}lcccccccc@{}}
  \toprule
  & \multicolumn{4}{c}{\textbf{Qwen2.5-3B-Instruct}} & \multicolumn{4}{c}{\textbf{Llama-3.2-3B-Instruct}} \\
  \cmidrule(lr){2-5} \cmidrule(lr){6-9}
  \textbf{Environment} & Baseline & \makecell{Baseline\\w/ Few Shots} & GRPO & CBRL GRPO & Baseline & \makecell{Baseline\\w/ Few Shots} & GRPO & CBRL GRPO \\
  \midrule
  ARC-1D            & $\phantom{0}4.67 \pm 0.33$ & $\phantom{0}2.00 \pm 0.58$ & $26.00 \pm 1.15$ & $\mathbf{30.67}$\,$\pm\,0.67$ & $\phantom{0}2.33 \pm 0.33$ & $\phantom{0}3.67 \pm 0.88$ & $17.00 \pm 0.58$ & $\mathbf{25.00}$\,$\pm\,0.58$ \\
  Manipulate Matrix & $\phantom{0}1.33 \pm 0.33$ & $\phantom{0}5.00 \pm 1.00$ & $\phantom{0}6.00 \pm 0.58$ & $\phantom{0}\mathbf{8.00}$\,$\pm\,0.58$ & $\phantom{0}0.33 \pm 0.33$ & $\phantom{0}2.67 \pm 0.88$ & $\phantom{0}3.33 \pm 0.88$ & $\phantom{0}\mathbf{8.33}$\,$\pm\,0.33$ \\
  Spell Backward    & $\phantom{0}4.33 \pm 1.33$ & $\phantom{0}4.00 \pm 0.58$ & $51.00 \pm 0.00$ & $\mathbf{52.67}$\,$\pm\,0.67$ & $20.00 \pm 3.46$ & $23.67 \pm 1.20$ & $95.67 \pm 0.33$ & $\mathbf{97.00}$\,$\pm\,0.58$ \\
  Word Sorting      & $\phantom{0}1.33 \pm 0.33$ & $\phantom{0}8.67 \pm 1.45$ & $53.33 \pm 0.67$ & $\mathbf{75.67}$\,$\pm\,1.20$ & $\phantom{0}8.67 \pm 1.67$ & $13.33 \pm 0.88$ & $80.00 \pm 0.58$ & $\mathbf{84.33}$\,$\pm\,0.88$ \\
  Puzzle24          & $\phantom{0}4.33 \pm 1.33$ & $\phantom{0}6.67 \pm 0.33$ & $48.00 \pm 1.00$ & $\mathbf{60.67}$\,$\pm\,0.88$ & $\phantom{0}2.67 \pm 0.88$ & $\phantom{0}4.00 \pm 2.08$ & $20.67 \pm 0.67$ & $\mathbf{22.00}$\,$\pm\,0.00$ \\
  \bottomrule
  \end{tabular}%
  }
\end{table*}
Table \ref{tab:sr_results_both_models_mean_se} compares baseline GRPO and CBRL-GRPO across all five Reasoning Gym environments for both Qwen2.5-3B-Instruct and Llama3.2-3B-Instruct. CBRL outperforms the baseline in every environment for both model families, demonstrating
that the method generalizes across architectures.

The magnitude of improvement varies by task and model. For Qwen2.5-3B, the largest gains appear on Word Sorting (+22.34\%) and Puzzle-24 (+12.67\%), while Llama3.2-3B shows the most improvement on ARC-1D (+8.00  points) and Manipulate Matrix (+5.00\%). This variation suggests that different models benefit from context bootstrapping in different ways, reflecting differences in their pretrained capabilities across task types. Notably, CBRL improves performance even on tasks where the baseline already achieves moderate accuracy, such as Spell Backward for Llama3.2-3B (95.67 → 97.00).

\begin{table}[h]
      \centering
      \caption{Performance comparison on Q programming language generation for QQwen-7B-Pretrain model. All results are mean over 5 samples and values represent percentages (\%). Valid Q Rate measures syntactically correct outputs; Avg.\ Test Pass and Success Rate measure functional correctness.}
      \label{tab:q_lang_results}
      \vspace{0.5em}
      \setlength{\tabcolsep}{6pt}
      \renewcommand{\arraystretch}{1.25}
      \begin{tabular}{@{}lccc@{}}
      \toprule
      \textbf{Method} & Valid Q (\%) & Avg.\ Pass (\%) & Success Rate (\%) \\
      \midrule
      Baseline              & 3.1 & 0.3 & 0.0 \\
      Baseline w/ Few Shots & 4.3 & 0.6 & 0.4 \\
      \midrule
      GRPO                  & \textbf{89.1} & 27.3 & 5.0 \\
      CBRL GRPO             & 80.9 & \textbf{43.0} & \textbf{26.3} \\
      \bottomrule
      \end{tabular}
  \end{table}

CBRL also demonstrates gains on Q programming problems (Table~\ref{tab:q_lang_results}). Here, it improves the overall pass@1 accuracy from 5.0\% to 26.3\%, indicating that context bootstrapping helps models navigate unfamiliar syntax and semantics more effectively. A closer look reveals that while standard GRPO largely optimizes for producing syntactically valid Q code, CBRL enables the model to go further, learning to actually apply the language's constructs to solve problems. Taken together, these results establish CBRL as a reliable, general-purpose enhancement for RLVR that provides consistent gains across models, task types, and domains.

\subsection{Analysis}

\paragraph{CBRL is algorithm agnostic.}

\begin{figure}[t]
  \centering
  \begin{minipage}[c]{0.48\linewidth}
    \centering
    \captionof{table}{Mean accuracy ($\pm$ standard error across three runs) of Qwen2.5-3B-Instruct trained with baseline RLOO and CBRL RLOO across five Reasoning Gym environments. Bold indicates the higher mean.}
    \label{tab:rloo_mean_se_qwen_simple}
    \vspace{0.5em}
    \setlength{\tabcolsep}{8pt}
    \renewcommand{\arraystretch}{1.3}
    \small
    \begin{tabular}{@{}lcc@{}}
    \toprule
    \textbf{Environment} & RLOO & CBRL RLOO \\
    \midrule
    ARC-1D            & $\mathbf{10.33}$\,$\pm\,0.33$ & $\phantom{0}8.00$\,$\pm\,0.58$ \\
    Word Sorting      & $20.33$\,$\pm\,1.45$ & $\mathbf{67.33}$\,$\pm\,1.20$ \\
    Puzzle24          & $23.00$\,$\pm\,0.00$ & $\mathbf{66.00}$\,$\pm\,1.73$ \\
    Spell Backward    & $63.67$\,$\pm\,1.20$ & $\mathbf{89.67}$\,$\pm\,0.33$ \\
    Manipulate Matrix & $\mathbf{8.67}$\,$\pm\,0.67$ & $\phantom{0}1.67$\,$\pm\,0.67$ \\
    \bottomrule
    \end{tabular}
  \end{minipage}
  \hfill
  \begin{minipage}[c]{0.48\linewidth}
    \centering
    \includegraphics[width=\linewidth]{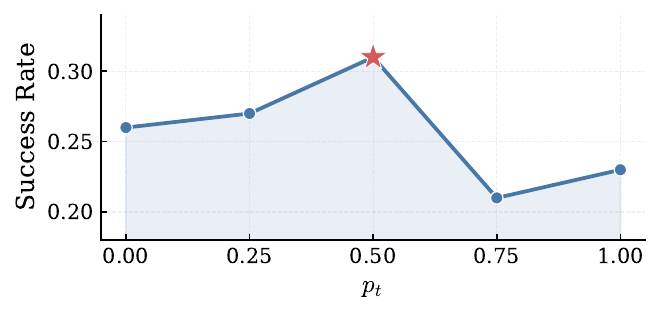}
    \captionof{figure}{Effect of initial injection probability $p_i$ on success rate for Qwen2.5-3B-Instruct trained with CBRL-GRPO on ARC-1D (500 steps). Performance peaks at $p_i = 0.5$; both extremes yield lower accuracy.}
    \label{fig:injection_prob}
  \end{minipage}
\end{figure}

To test whether CBRL is specific to GRPO or transfers to other policy-gradient methods, we apply it to REINFORCE Leave-One-Out (RLOO)~\citep{DBLP:conf/acl/AhmadianCGFKPUH24}. Table~\ref{tab:rloo_mean_se_qwen_simple} reports mean accuracy ($\pm$ standard error across three runs) for Qwen2.5-3B-Instruct trained with baseline RLOO and CBRL-RLOO across five Reasoning Gym environments.

CBRL improves performance on three of the five tasks, with large gains on Word Sorting (+47 points), Puzzle24 (+43 points), and Spell Backward (+26 points). In contrast, it reduces performance on ARC-1D (-2.3 points) and Manipulate Matrix (-7.0 points). Notably, the gains under RLOO are larger than those observed with GRPO (Table~\ref{tab:sr_results_both_models_mean_se}), which may stem from RLOO’s higher-variance gradient estimates, which can make early learning less stable and increase the value of context-based guidance.



Overall, these results suggest that CBRL is not tied to a particular algorithm. It can be combined with different policy-gradient methods without modification and yields substantial improvements on several tasks. The failures on ARC-1D and Manipulate Matrix indicate that the effectiveness of CBRL depends on the match between selected context and task structure; developing task-aware context selection strategies is an important direction for future work.

\paragraph{CBRL improves exploration efficiency and retains gains after few-shots are annealed.}

\begin{figure*}[h]
    \begin{center}
        \includegraphics[width=0.95\linewidth]{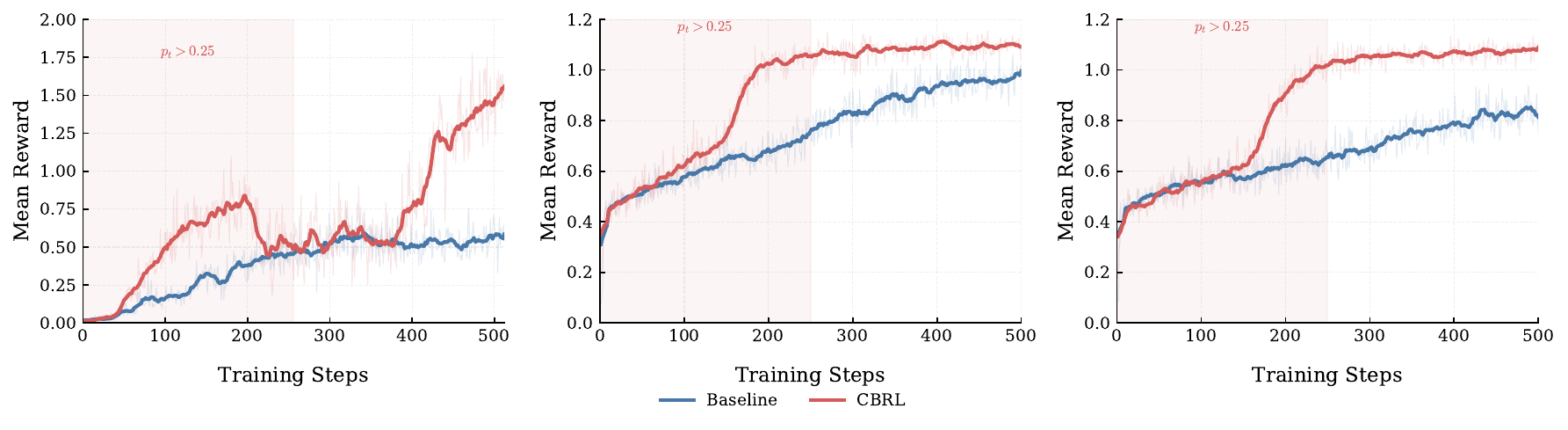} 
    \end{center}
    \vspace{-0.5em}
    \caption{Training reward curves for CBRL and baseline across three settings: Q Programming with GRPO (left), Word Sorting with GRPO (center), and Word Sorting with RLOO (right). Shaded regions indicate $p_i > 0.25$ (high injection). CBRL achieves higher early reward by guiding the model toward successful rollouts, bootstrapping the learning process. The advantage persists after injection stops, demonstrating that CBRL addresses exploration inefficiency without creating long-term dependence on demonstrations.}
    \label{fig:training_curves}

\end{figure*}

Figure~\ref{fig:training_curves} illustrates training reward curves across three settings: Q Programming with GRPO, Word Sorting with GRPO, and Word Sorting with RLOO. The shaded regions indicate steps where $p_i > 0.25$, when few-shot demonstrations are injected most frequently.

During the early injection phase ($p_i > 0.25$), CBRL-trained models achieve higher reward than baselines across all three settings. The injected examples guide them toward successful rollouts that they may not discover independently. These early successes provide the reward signal needed to bootstrap learning. Once the model has learned to generate successful rollouts on its own, it no longer requires demonstrations. As $p_i$ anneals toward zero, CBRL maintains its advantage. The fact that this pattern holds across both GRPO and RLOO, and across reasoning and code generation domains, reinforces that the exploration benefit is not tied to a particular algorithm or task type.

The baseline, by contrast, must discover successful strategies through random exploration alone. The model may spend several steps generating failed rollouts before stumbling upon a successful one. CBRL sidesteps this bottleneck by providing early guidance, enabling the model to quickly learn new reasoning patterns or domain-specific details.

\paragraph{Effect of initial injection probability on CBRL training.}

Figure \ref{fig:injection_prob} shows the success rate of Qwen2.5-3B-Instruct on ARC-1D environment after 500 training steps as a function of the initial injection probability $p_i$. Performance follows an inverted-V pattern. Success rate rises from 0.26 at $p_i{=}0.0$ (no context injection) to a peak of 0.31 at $p_i{=}0.5$, then drops to 0.23 at $p_i{=}1.0$. This suggests that moderate context bootstrapping provides a useful learning signal during early training, while injecting examples too frequently prevents the model from learning to reason independently, ultimately hurting final performance.

\begin{figure*}[t]
  \centering

  \definecolor{boxblue}{RGB}{217,230,245}
  \definecolor{boxgreen}{RGB}{216,237,204}
  \definecolor{boxgray}{RGB}{240,240,240}
  \definecolor{borderblue}{RGB}{120,150,190}
  \definecolor{borderred}{RGB}{190,80,80}
  \definecolor{bordergreen}{RGB}{100,160,90}

  \tcbset{
    qualstyle/.style={
      enhanced,
      fonttitle=\bfseries\small,
      fontupper=\small,
      boxrule=0.4pt,
      arc=2pt,
      top=4pt, bottom=4pt, left=5pt, right=5pt,
      equal height group=threeblocks,
      before upper={\setlength{\parskip}{4pt}},
    }
  }

  \begin{minipage}[t]{0.32\linewidth}\vspace{0pt}
  \begin{tcolorbox}[qualstyle,
    colback=boxblue!30, colframe=borderblue,
    colbacktitle=borderblue, coltitle=white,
    title=Baseline]
  \textit{\textbf{Think:}}\\
  The words provided are: violates, yes, already, completing, pages, duty, his, EXPRESS, duly. We need to sort these words in ascending order using ASCII/Unicode ordering. The sorting will consider letter cases and special characters. Let's sort the words.

  \textit{\textbf{Answer:}}\\
  EXPRESS, already, completing, duty, his, violates, pages, duly, yes
  \end{tcolorbox}
  \end{minipage}\hfill
  \begin{minipage}[t]{0.32\linewidth}\vspace{0pt}
  \begin{tcolorbox}[qualstyle,
    colback=boxgreen!30, colframe=bordergreen,
    colbacktitle=bordergreen, coltitle=white,
    title=CBRL]
  \textit{\textbf{Think:}}\\
  The words need to be sorted in ascending ASCII order. We start by comparing the ASCII values of the first characters of the words.\\
  The ASCII value of \texttt{'v'} in ``violates'' is 118.\\
  The ASCII value of \texttt{'y'} in ``yes'' is 121. \\
  The ASCII value of \texttt{'a'} in "already" is 97.\\
  \dots \\
  Sorting these values in ascending order gives: 69, 97, 99, 100, 100, 104, 112, 118, 121.\\
  Mapping these back to characters gives: \texttt{'E'}, \texttt{'a'}, \texttt{'c'}, \texttt{'d'}, \texttt{'d'}, \texttt{'h'}, \texttt{'p'}, \texttt{'v'}, \texttt{'y'}. \\
  \dots

  \textit{\textbf{Answer:}}\\
  EXPRESS, already, completing, duly, duty, his, pages, violates, yes
  \end{tcolorbox}
  \end{minipage}\hfill
  \begin{minipage}[t]{0.32\linewidth}\vspace{0pt}
  \begin{tcolorbox}[qualstyle,
    colback=boxgray, colframe=gray!60,
    colbacktitle=gray!60, coltitle=white,
    title=Few-Shot Example]
  \textit{\textbf{Think:}}\\
  Use ASCII/Unicode lexicographic ordering: compare strings by code points left to right;
  \dots \\
  Next is ``Transform'' because it begins with uppercase \texttt{'T'}~(84), while all remaining words start with lowercase letters ($\ge 97$). \\
  \dots \\
  Between ``following'' and ``forgetful,'' both start with ``fo,'' but \texttt{'l'}~(108) $<$ \texttt{'r'}~(114), so ``following'' precedes ``forgetful'';
  ``turning'' (\texttt{t}=116) comes last. \\
  \dots

  \textit{\textbf{Answer:}}\\
  ``799'', ``Transform'', ``business'', ``days'', ``dead'', ``following'', ``forgetful'', ``turning''
  \end{tcolorbox}
  \end{minipage}

  \caption{Qualitative comparison of model outputs on the task: ``Sort the words \texttt{violates, yes, already, completing, pages, duty, his, EXPRESS, duly} in ascending ASCII order.''
  {\setlength{\fboxsep}{1pt}\colorbox{boxblue}{Baseline}} (left) offers a superficial explanation and produces an incorrect ordering.
  {\setlength{\fboxsep}{1pt}\colorbox{boxgreen}{CBRL}} (center) explicitly reasons about ASCII values (e.g., \texttt{'E'}\,=\,69, \texttt{'a'}\,=\,97), systematically compares characters, and arrives at the correct answer.
  The \textbf{few-shot example} (right) shows the step-by-step ASCII reasoning pattern used during CBRL training.}
  \label{fig:qualitative}
\end{figure*}
\paragraph{Qualitative Result from CBRL vs. Baseline Training}

To illustrate the behavioral differences induced by CBRL, we present representative model outputs on a Word Sorting instance in Figure~\ref{fig:qualitative}. The task requires sorting a list of words in ascending ASCII order, which demands explicit knowledge of character code points and careful left-to-right comparison.

The baseline model (Figure~\ref{fig:qualitative}, left) acknowledges the need for ASCII ordering but fails to engage in systematic character-level comparison. Its response remains at a surface level, mentioning ``letter cases and special characters" without computing any code points, and produces an incorrect ordering.

In contrast, the CBRL-trained model (Figure~\ref{fig:qualitative}, center) exhibits structured, step-by-step reasoning that closely mirrors the few-shot exemplars used during training (Figure~\ref{fig:qualitative}, right). It explicitly retrieves ASCII values for initial characters (e.g., \texttt{'v'}$\,=\,$118, \texttt{'y'}$\,=\,$121, \texttt{'a'}$\,=\,$97), sorts the numerical values, maps them back to the corresponding words, and arrives at the correct answer. 
\section{Related Work}

\subsection{Reinforcement Learning from Verifiable Rewards}

Reinforcement Learning from Verifiable Rewards (RLVR)~\citep{tulu3} has emerged as the dominant paradigm for post-training large reasoning models (LRMs). Unlike Reinforcement Learning from Human Feedback (RLHF) approaches that rely on learned reward models, RLVR leverages deterministic verifiers providing binary feedback based on answer correctness. This makes it particularly effective for mathematical reasoning and code generation~\citep{deepseekmath, deepseekr1}. Large reasoning models such as DeepSeek-R1~\citep{deepseekr1}, QwQ~\citep{qwen} and Kimi-1.5~\citep{kimik15} have demonstrated that RLVR enables superior generalization and robustness compared to SFT alone~\citep{chusftrlvr}.

Group Relative Policy Optimization (GRPO)~\citep{deepseekmath} has become the de facto algorithm for RLVR training, eliminating the need for a separate value model by computing advantages relative to sampled completions within each batch. DeepSeek-R1~\citep{deepseekr1} demonstrated that GRPO-based training can elicit sophisticated reasoning behaviors, including the ``aha moment'' where models spontaneously develop self-reflection capabilities. Subsequent work has systematically investigated zero RL training, training base models directly with RLVR without SFT warm-up~\citep{simplerlzoo, dapo, drgrpo}.

However, a critical failure mode for RLVR is the \emph{exploration inefficiency problem}: when all trajectories in a sample group are incorrect, the group advantage collapses to zero, yielding no gradient for policy updates. This reward sparsity is particularly severe when models have not been exposed to a certain domain or reasoning style.

\subsection{Addressing the Exploration Inefficiency Problem in RLVR}

 Recent approaches address this challenge through two main strategies. The first is \textbf{mixed-policy training}, which interleaves RL with SFT or replaces portions of on-policy rollouts with high-quality off-policy trajectories~\citep{luffy, onpolrloffpolexp}. LUFFY~\citep{luffy} leverages off-policy reasoning traces to transcend cognitive constraints while preserving self-driven exploration through regularized importance sampling. 
 
 The second strategy employs \textbf{partial supervision or hints}, providing segments of ground-truth solutions to rescue failed rollouts and guide exploration along correct trajectories~\citep{hint, ghpo, bread}. While these approaches effectively improve rollout efficiency, over-reliance on off-policy data can misguide policy updates toward non-generalizable solution paths.

Curriculum-based approaches offer another direction. E2H Reasoner~\citep{e2h} schedules tasks from easy to hard, showing that fading out easy tasks prevents overfitting. Absolute Zero Reasoner~\citep{azr} self-evolves its curriculum using a code executor to validate tasks and verify answers.

CBRL combines the benefits of these strategies while avoiding their limitations. Unlike mixed-policy methods, CBRL remains fully on-policy. In-context examples serve as context rather than trajectories to imitate, preserving the exploration dynamics that enable robust generalization. Unlike partial supervision approaches, CBRL does not intervene mid-rollout; instead, it leverages native in-context learning to bootstrap exploration from the start. 

Furthermore, unlike curriculum methods requiring task difficulty estimation or self-generated problems, CBRL operates on fixed distributions with a simple annealing schedule, creating an implicit curriculum from guided to independent reasoning without external infrastructure. 
\section{Conclusion}

We introduced Context Bootstrapped Reinforcement Learning (CBRL), a method that addresses exploration inefficiency in RLVR by stochastically injecting few-shot demonstrations during training. The injection probability follows a curriculum
that starts high to bootstrap early exploration and anneals to zero, forcing the model to internalize demonstrated reasoning patterns rather than depend on them at inference time.

Our experiments demonstrate that CBRL consistently improves performance across two model families, five reasoning tasks, and two policy gradient algorithms. The method yields gains ranging from +1.3\% to +22.3\% on Reasoning Gym tasks and
improves the success rate from 5.0\% to 26.3\% on Q programming, a domain-specific language with unconventional syntax. Training curves confirm that CBRL accelerates early learning and that gains persist after demonstrations are
removed, indicating durable capability acquisition rather than superficial reliance on in-context examples.

CBRL demonstrates that the in-context learning capabilities of LLMs can serve as a bridge for reinforcement learning, providing temporary scaffolding that bootstraps exploration without compromising the model's ability to generalize
independently. We hope this work encourages further investigation into leveraging non-parametric, context-based strategies to address fundamental challenges in parametric post-training.
\section{Future Work}

We see several promising directions for future work. First, the injection schedule could be made adaptive, adjusting $p_i$ based on reward trends or success rates rather than following a fixed linear decay. Second, developing principled methods for constructing and selecting demonstrations remains an open challenge; learned retrieval mechanisms could automate this process while improving alignment between examples and training instances.

Third, extending CBRL to longer-horizon settings, including multi-step reasoning chains and agentic workflows, could address compounding exploration difficulties where early missteps preclude the discovery of correct solution paths. Finally, investigating the interaction between CBRL and model scale, as well as combining CBRL with complementary approaches such as off-policy learning, may yield further gains.

\section*{Acknowledgment}
This work used DeltaAI at the National Center for Supercomputing Applications (NCSA) through allocation CIS260052 from the Advanced Cyberinfrastructure Coordination Ecosystem: Services \& Support (ACCESS) program, which is supported by U.S. National Science Foundation grants \#2138259, \#2138286, \#2138307, \#2137603, and \#2138296.




\bibliography{main}

@article{deepseekmath,
  author       = {Zhihong Shao and
                  Peiyi Wang and
                  Qihao Zhu and
                  Runxin Xu and
                  Junxiao Song and
                  Mingchuan Zhang and
                  Y. K. Li and
                  Y. Wu and
                  Daya Guo},
  title        = {DeepSeekMath: Pushing the Limits of Mathematical Reasoning in Open
                  Language Models},
  journal      = {CoRR},
  volume       = {abs/2402.03300},
  year         = {2024},
  url          = {https://doi.org/10.48550/arXiv.2402.03300},
  doi          = {10.48550/ARXIV.2402.03300},
  eprinttype    = {arXiv},
  eprint       = {2402.03300},
  bibsource    = {dblp computer science bibliography, https://dblp.org}
}

@article{deepseekr1,
  author       = {DeepSeek{-}AI},
  title        = {DeepSeek-R1: Incentivizing Reasoning Capability in LLMs via Reinforcement
                  Learning},
  journal      = {CoRR},
  volume       = {abs/2501.12948},
  year         = {2025},
  url          = {https://doi.org/10.48550/arXiv.2501.12948},
  doi          = {10.48550/ARXIV.2501.12948},
  eprinttype    = {arXiv},
  eprint       = {2501.12948},
  bibsource    = {dblp computer science bibliography, https://dblp.org}
}

@article{tulu3,
  author       = {Nathan Lambert and
                  Jacob Morrison and
                  Valentina Pyatkin and
                  Shengyi Huang and
                  Hamish Ivison and
                  Faeze Brahman and
                  Lester James V. Miranda and
                  Alisa Liu and
                  Nouha Dziri and
                  Shane Lyu and
                  Yuling Gu and
                  Saumya Malik and
                  Victoria Graf and
                  Jena D. Hwang and
                  Jiangjiang Yang and
                  Ronan Le Bras and
                  Oyvind Tafjord and
                  Chris Wilhelm and
                  Luca Soldaini and
                  Noah A. Smith and
                  Yizhong Wang and
                  Pradeep Dasigi and
                  Hannaneh Hajishirzi},
  title        = {T{\"{U}}LU 3: Pushing Frontiers in Open Language Model Post-Training},
  journal      = {CoRR},
  volume       = {abs/2411.15124},
  year         = {2024},
  url          = {https://doi.org/10.48550/arXiv.2411.15124},
  doi          = {10.48550/ARXIV.2411.15124},
  eprinttype    = {arXiv},
  eprint       = {2411.15124},
  bibsource    = {dblp computer science bibliography, https://dblp.org}
}

@article{kimik15,
  author       = {Kimi Team and
                  Angang Du and
                  Bofei Gao and
                  Bowei Xing and
                  Changjiu Jiang and
                  Cheng Chen and
                  Cheng Li and
                  Chenjun Xiao and
                  Chenzhuang Du and
                  Chonghua Liao and
                  Chuning Tang and
                  Congcong Wang and
                  Dehao Zhang and
                  Enming Yuan and
                  Enzhe Lu and
                  Fengxiang Tang and
                  Flood Sung and
                  Guangda Wei and
                  Guokun Lai and
                  Haiqing Guo and
                  Han Zhu and
                  Hao Ding and
                  Hao Hu and
                  Hao Yang and
                  Hao Zhang and
                  Haotian Yao and
                  Haotian Zhao and
                  Haoyu Lu and
                  Haoze Li and
                  Haozhen Yu and
                  Hongcheng Gao and
                  Huabin Zheng and
                  Huan Yuan and
                  Jia Chen and
                  Jianhang Guo and
                  Jianlin Su and
                  Jianzhou Wang and
                  Jie Zhao and
                  Jin Zhang and
                  Jingyuan Liu and
                  Junjie Yan and
                  Junyan Wu and
                  Lidong Shi and
                  Ling Ye and
                  Longhui Yu and
                  Mengnan Dong and
                  Neo Zhang and
                  Ningchen Ma and
                  Qiwei Pan and
                  Qucheng Gong and
                  Shaowei Liu and
                  Shengling Ma and
                  Shupeng Wei and
                  Sihan Cao and
                  Siying Huang and
                  Tao Jiang and
                  Weihao Gao and
                  Weimin Xiong and
                  Weiran He and
                  Weixiao Huang and
                  Wenhao Wu and
                  Wenyang He and
                  Xianghui Wei and
                  Xianqing Jia and
                  Xingzhe Wu and
                  Xinran Xu and
                  Xinxing Zu and
                  Xinyu Zhou and
                  Xuehai Pan and
                  Y. Charles and
                  Yang Li and
                  Yangyang Hu and
                  Yangyang Liu and
                  Yanru Chen and
                  Yejie Wang and
                  Yibo Liu and
                  Yidao Qin and
                  Yifeng Liu and
                  Ying Yang and
                  Yiping Bao and
                  Yulun Du and
                  Yuxin Wu and
                  Yuzhi Wang and
                  Zaida Zhou and
                  Zhaoji Wang and
                  Zhaowei Li and
                  Zhen Zhu and
                  Zheng Zhang and
                  Zhexu Wang and
                  Zhilin Yang and
                  Zhiqi Huang and
                  Zihao Huang and
                  Ziyao Xu and
                  Zonghan Yang},
  title        = {Kimi k1.5: Scaling Reinforcement Learning with LLMs},
  journal      = {CoRR},
  volume       = {abs/2501.12599},
  year         = {2025},
  url          = {https://doi.org/10.48550/arXiv.2501.12599},
  doi          = {10.48550/ARXIV.2501.12599},
  eprinttype    = {arXiv},
  eprint       = {2501.12599},
  bibsource    = {dblp computer science bibliography, https://dblp.org}
}

@inproceedings{chusftrlvr,
  author       = {Tianzhe Chu and
                  Yuexiang Zhai and
                  Jihan Yang and
                  Shengbang Tong and
                  Saining Xie and
                  Dale Schuurmans and
                  Quoc V. Le and
                  Sergey Levine and
                  Yi Ma},
  title        = {{SFT} Memorizes, {RL} Generalizes: {A} Comparative Study of Foundation
                  Model Post-training},
  booktitle    = {Forty-second International Conference on Machine Learning, {ICML}
                  2025, Vancouver, BC, Canada, July 13-19, 2025},
  publisher    = {OpenReview.net},
  year         = {2025},
  url          = {https://openreview.net/forum?id=dYur3yabMj},
  bibsource    = {dblp computer science bibliography, https://dblp.org}
}

@article{simplerlzoo,
  author       = {Weihao Zeng and
                  Yuzhen Huang and
                  Qian Liu and
                  Wei Liu and
                  Keqing He and
                  Zejun Ma and
                  Junxian He},
  title        = {SimpleRL-Zoo: Investigating and Taming Zero Reinforcement Learning
                  for Open Base Models in the Wild},
  journal      = {CoRR},
  volume       = {abs/2503.18892},
  year         = {2025},
  url          = {https://doi.org/10.48550/arXiv.2503.18892},
  doi          = {10.48550/ARXIV.2503.18892},
  eprinttype    = {arXiv},
  eprint       = {2503.18892},
  bibsource    = {dblp computer science bibliography, https://dblp.org}
}

@article{dapo,
  author       = {Qiying Yu and
                  Zheng Zhang and
                  Ruofei Zhu and
                  Yufeng Yuan and
                  Xiaochen Zuo and
                  Yu Yue and
                  Tiantian Fan and
                  Gaohong Liu and
                  Lingjun Liu and
                  Xin Liu and
                  Haibin Lin and
                  Zhiqi Lin and
                  Bole Ma and
                  Guangming Sheng and
                  Yuxuan Tong and
                  Chi Zhang and
                  Mofan Zhang and
                  Wang Zhang and
                  Hang Zhu and
                  Jinhua Zhu and
                  Jiaze Chen and
                  Jiangjie Chen and
                  Chengyi Wang and
                  Hongli Yu and
                  Weinan Dai and
                  Yuxuan Song and
                  Xiangpeng Wei and
                  Hao Zhou and
                  Jingjing Liu and
                  Wei{-}Ying Ma and
                  Ya{-}Qin Zhang and
                  Lin Yan and
                  Mu Qiao and
                  Yonghui Wu and
                  Mingxuan Wang},
  title        = {{DAPO:} An Open-Source {LLM} Reinforcement Learning System at Scale},
  journal      = {CoRR},
  volume       = {abs/2503.14476},
  year         = {2025},
  url          = {https://doi.org/10.48550/arXiv.2503.14476},
  doi          = {10.48550/ARXIV.2503.14476},
  eprinttype    = {arXiv},
  eprint       = {2503.14476},
  bibsource    = {dblp computer science bibliography, https://dblp.org}
}

@article{drgrpo,
  author       = {Zichen Liu and
                  Changyu Chen and
                  Wenjun Li and
                  Penghui Qi and
                  Tianyu Pang and
                  Chao Du and
                  Wee Sun Lee and
                  Min Lin},
  title        = {Understanding R1-Zero-Like Training: {A} Critical Perspective},
  journal      = {CoRR},
  volume       = {abs/2503.20783},
  year         = {2025},
  url          = {https://doi.org/10.48550/arXiv.2503.20783},
  doi          = {10.48550/ARXIV.2503.20783},
  eprinttype    = {arXiv},
  eprint       = {2503.20783},
  bibsource    = {dblp computer science bibliography, https://dblp.org}
}

@article{yue2025limit,
  author       = {Yang Yue and
                  Zhiqi Chen and
                  Rui Lu and
                  Andrew Zhao and
                  Zhaokai Wang and
                  Yang Yue and
                  Shiji Song and
                  Gao Huang},
  title        = {Does Reinforcement Learning Really Incentivize Reasoning Capacity
                  in LLMs Beyond the Base Model?},
  journal      = {CoRR},
  volume       = {abs/2504.13837},
  year         = {2025},
  url          = {https://doi.org/10.48550/arXiv.2504.13837},
  doi          = {10.48550/ARXIV.2504.13837},
  eprinttype    = {arXiv},
  eprint       = {2504.13837},
  bibsource    = {dblp computer science bibliography, https://dblp.org}
}

@article{luffy,
  author       = {Jianhao Yan and
                  Yafu Li and
                  Zican Hu and
                  Zhi Wang and
                  Ganqu Cui and
                  Xiaoye Qu and
                  Yu Cheng and
                  Yue Zhang},
  title        = {Learning to Reason under Off-Policy Guidance},
  journal      = {CoRR},
  volume       = {abs/2504.14945},
  year         = {2025},
  url          = {https://doi.org/10.48550/arXiv.2504.14945},
  doi          = {10.48550/ARXIV.2504.14945},
  eprinttype    = {arXiv},
  eprint       = {2504.14945},
  bibsource    = {dblp computer science bibliography, https://dblp.org}
}

@article{onpolrloffpolexp,
  author       = {Wenhao Zhang and
                  Yuexiang Xie and
                  Yuchang Sun and
                  Yanxi Chen and
                  Guoyin Wang and
                  Yaliang Li and
                  Bolin Ding and
                  Jingren Zhou},
  title        = {On-Policy {RL} Meets Off-Policy Experts: Harmonizing Supervised Fine-Tuning
                  and Reinforcement Learning via Dynamic Weighting},
  journal      = {CoRR},
  volume       = {abs/2508.11408},
  year         = {2025},
  url          = {https://doi.org/10.48550/arXiv.2508.11408},
  doi          = {10.48550/ARXIV.2508.11408},
  eprinttype    = {arXiv},
  eprint       = {2508.11408},
  bibsource    = {dblp computer science bibliography, https://dblp.org}
}

@article{hint,
  author       = {Xinyi Wang and
                  Jinyi Han and
                  Zishang Jiang and
                  Tingyun Li and
                  Jiaqing Liang and
                  Sihang Jiang and
                  Zhaoqian Dai and
                  Shuguang Ma and
                  Fei Yu and
                  Yanghua Xiao},
  title        = {{HINT:} Helping Ineffective Rollouts Navigate Towards Effectiveness},
  journal      = {CoRR},
  volume       = {abs/2510.09388},
  year         = {2025},
  url          = {https://doi.org/10.48550/arXiv.2510.09388},
  doi          = {10.48550/ARXIV.2510.09388},
  eprinttype    = {arXiv},
  eprint       = {2510.09388},
  bibsource    = {dblp computer science bibliography, https://dblp.org}
}

@article{ghpo,
  author       = {Ziru Liu and
                  Cheng Gong and
                  Xinyu Fu and
                  Yaofang Liu and
                  Ran Chen and
                  Shoubo Hu and
                  Suiyun Zhang and
                  Rui Liu and
                  Qingfu Zhang and
                  Dandan Tu},
  title        = {{GHPO:} Adaptive Guidance for Stable and Efficient {LLM} Reinforcement
                  Learning},
  journal      = {CoRR},
  volume       = {abs/2507.10628},
  year         = {2025},
  url          = {https://doi.org/10.48550/arXiv.2507.10628},
  doi          = {10.48550/ARXIV.2507.10628},
  eprinttype    = {arXiv},
  eprint       = {2507.10628},
  bibsource    = {dblp computer science bibliography, https://dblp.org}
}

@article{bread,
  author       = {Xuechen Zhang and
                  Zijian Huang and
                  Yingcong Li and
                  Chenshun Ni and
                  Jiasi Chen and
                  Samet Oymak},
  title        = {{BREAD:} Branched Rollouts from Expert Anchors Bridge {SFT} {\&}
                  {RL} for Reasoning},
  journal      = {CoRR},
  volume       = {abs/2506.17211},
  year         = {2025},
  url          = {https://doi.org/10.48550/arXiv.2506.17211},
  doi          = {10.48550/ARXIV.2506.17211},
  eprinttype    = {arXiv},
  eprint       = {2506.17211},
  bibsource    = {dblp computer science bibliography, https://dblp.org}
}

@article{e2h,
  author       = {Shubham Parashar and
                  Shurui Gui and
                  Xiner Li and
                  Hongyi Ling and
                  Sushil Vemuri and
                  Blake Olson and
                  Eric Li and
                  Yu Zhang and
                  James Caverlee and
                  Dileep Kalathil and
                  Shuiwang Ji},
  title        = {Curriculum Reinforcement Learning from Easy to Hard Tasks Improves
                  {LLM} Reasoning},
  journal      = {CoRR},
  volume       = {abs/2506.06632},
  year         = {2025},
  url          = {https://doi.org/10.48550/arXiv.2506.06632},
  doi          = {10.48550/ARXIV.2506.06632},
  eprinttype    = {arXiv},
  eprint       = {2506.06632},
  bibsource    = {dblp computer science bibliography, https://dblp.org}
}

@article{azr,
  author       = {Andrew Zhao and
                  Yiran Wu and
                  Yang Yue and
                  Tong Wu and
                  Quentin Xu and
                  Yang Yue and
                  Matthieu Lin and
                  Shenzhi Wang and
                  Qingyun Wu and
                  Zilong Zheng and
                  Gao Huang},
  title        = {Absolute Zero: Reinforced Self-play Reasoning with Zero Data},
  journal      = {CoRR},
  volume       = {abs/2505.03335},
  year         = {2025},
  url          = {https://doi.org/10.48550/arXiv.2505.03335},
  doi          = {10.48550/ARXIV.2505.03335},
  eprinttype    = {arXiv},
  eprint       = {2505.03335},
  bibsource    = {dblp computer science bibliography, https://dblp.org}
}

@article{qwen,
  author       = {An Yang and
                  Baosong Yang and
                  Beichen Zhang and
                  Binyuan Hui and
                  Bo Zheng and
                  Bowen Yu and
                  Chengyuan Li and
                  Dayiheng Liu and
                  Fei Huang and
                  Haoran Wei and
                  Huan Lin and
                  Jian Yang and
                  Jianhong Tu and
                  Jianwei Zhang and
                  Jianxin Yang and
                  Jiaxi Yang and
                  Jingren Zhou and
                  Junyang Lin and
                  Kai Dang and
                  Keming Lu and
                  Keqin Bao and
                  Kexin Yang and
                  Le Yu and
                  Mei Li and
                  Mingfeng Xue and
                  Pei Zhang and
                  Qin Zhu and
                  Rui Men and
                  Runji Lin and
                  Tianhao Li and
                  Tingyu Xia and
                  Xingzhang Ren and
                  Xuancheng Ren and
                  Yang Fan and
                  Yang Su and
                  Yichang Zhang and
                  Yu Wan and
                  Yuqiong Liu and
                  Zeyu Cui and
                  Zhenru Zhang and
                  Zihan Qiu},
  title        = {Qwen2.5 Technical Report},
  journal      = {CoRR},
  volume       = {abs/2412.15115},
  year         = {2024},
  url          = {https://doi.org/10.48550/arXiv.2412.15115},
  doi          = {10.48550/ARXIV.2412.15115},
  eprinttype    = {arXiv},
  eprint       = {2412.15115},
  bibsource    = {dblp computer science bibliography, https://dblp.org}
}

@article{llama,
  author       = {Llama Team},
  title        = {The Llama 3 Herd of Models},
  journal      = {CoRR},
  volume       = {abs/2407.21783},
  year         = {2024},
  url          = {https://doi.org/10.48550/arXiv.2407.21783},
  doi          = {10.48550/ARXIV.2407.21783},
  eprinttype    = {arXiv},
  eprint       = {2407.21783},
  bibsource    = {dblp computer science bibliography, https://dblp.org}
}

@article{DBLP:journals/corr/abs-2505-24760,
  author       = {Zafir Stojanovski and
                  Oliver Stanley and
                  Joe Sharratt and
                  Richard Jones and
                  Abdulhakeem Adefioye and
                  Jean Kaddour and
                  Andreas K{\"{o}}pf},
  title        = {{REASONING} {GYM:} Reasoning Environments for Reinforcement Learning
                  with Verifiable Rewards},
  journal      = {CoRR},
  volume       = {abs/2505.24760},
  year         = {2025},
  url          = {https://doi.org/10.48550/arXiv.2505.24760},
  doi          = {10.48550/ARXIV.2505.24760},
  eprinttype    = {arXiv},
  eprint       = {2505.24760},
  bibsource    = {dblp computer science bibliography, https://dblp.org}
}

@article{DBLP:journals/corr/abs-2508-06813,
  author       = {Brendan R. Hogan and
                  Will Brown and
                  Adel Boyarsky and
                  Anderson Schneider and
                  Yuriy Nevmyvaka},
  title        = {Technical Report: Full-Stack Fine-Tuning for the {Q} Programming Language},
  journal      = {CoRR},
  volume       = {abs/2508.06813},
  year         = {2025},
  url          = {https://doi.org/10.48550/arXiv.2508.06813},
  doi          = {10.48550/ARXIV.2508.06813},
  eprinttype    = {arXiv},
  eprint       = {2508.06813},
  bibsource    = {dblp computer science bibliography, https://dblp.org}
}

@inproceedings{DBLP:conf/acl/AhmadianCGFKPUH24,
  author       = {Arash Ahmadian and
                  Chris Cremer and
                  Matthias Gall{\'{e}} and
                  Marzieh Fadaee and
                  Julia Kreutzer and
                  Olivier Pietquin and
                  Ahmet {\"{U}}st{\"{u}}n and
                  Sara Hooker},
  editor       = {Lun{-}Wei Ku and
                  Andre Martins and
                  Vivek Srikumar},
  title        = {Back to Basics: Revisiting REINFORCE-Style Optimization for Learning
                  from Human Feedback in LLMs},
  booktitle    = {Proceedings of the 62nd Annual Meeting of the Association for Computational
                  Linguistics (Volume 1: Long Papers), {ACL} 2024, Bangkok, Thailand,
                  August 11-16, 2024},
  pages        = {12248--12267},
  publisher    = {Association for Computational Linguistics},
  year         = {2024},
  url          = {https://doi.org/10.18653/v1/2024.acl-long.662},
  doi          = {10.18653/V1/2024.ACL-LONG.662},
  bibsource    = {dblp computer science bibliography, https://dblp.org}
}

@article{DBLP:journals/corr/abs-2505-11831,
  author       = {Fran{\c{c}}ois Chollet and
                  Mike Knoop and
                  Gregory Kamradt and
                  Bryan Landers and
                  Henry Pinkard},
  title        = {{ARC-AGI-2:} {A} New Challenge for Frontier {AI} Reasoning Systems},
  journal      = {CoRR},
  volume       = {abs/2505.11831},
  year         = {2025},
  url          = {https://doi.org/10.48550/arXiv.2505.11831},
  doi          = {10.48550/ARXIV.2505.11831},
  eprinttype    = {arXiv},
  eprint       = {2505.11831},
  bibsource    = {dblp computer science bibliography, https://dblp.org}
}

@article{DBLP:journals/corr/abs-2506-10947,
  author       = {Rulin Shao and
                  Shuyue Stella Li and
                  Rui Xin and
                  Scott Geng and
                  Yiping Wang and
                  Sewoong Oh and
                  Simon Shaolei Du and
                  Nathan Lambert and
                  Sewon Min and
                  Ranjay Krishna and
                  Yulia Tsvetkov and
                  Hannaneh Hajishirzi and
                  Pang Wei Koh and
                  Luke Zettlemoyer},
  title        = {Spurious Rewards: Rethinking Training Signals in {RLVR}},
  journal      = {CoRR},
  volume       = {abs/2506.10947},
  year         = {2025},
  url          = {https://doi.org/10.48550/arXiv.2506.10947},
  doi          = {10.48550/ARXIV.2506.10947},
  eprinttype    = {arXiv},
  eprint       = {2506.10947},
  bibsource    = {dblp computer science bibliography, https://dblp.org}
}

@misc{liu2025evocotovercomingexplorationbottleneck,
      title={EvoCoT: Overcoming the Exploration Bottleneck in Reinforcement Learning}, 
      author={Huanyu Liu and Jia Li and Chang Yu and Taozhi Chen and Yihong Dong and Lecheng Wang and XiaoLong Hu and Ge Li},
      year={2025},
      eprint={2508.07809},
      archivePrefix={arXiv},
      primaryClass={cs.LG},
      url={https://arxiv.org/abs/2508.07809}, 
}

@inproceedings{fewshot,
 author = {Brown, Tom and Mann, Benjamin and Ryder, Nick and Subbiah, Melanie and Kaplan, Jared D and Dhariwal, Prafulla and Neelakantan, Arvind and Shyam, Pranav and Sastry, Girish and Askell, Amanda and Agarwal, Sandhini and Herbert-Voss, Ariel and Krueger, Gretchen and Henighan, Tom and Child, Rewon and Ramesh, Aditya and Ziegler, Daniel and Wu, Jeffrey and Winter, Clemens and Hesse, Chris and Chen, Mark and Sigler, Eric and Litwin, Mateusz and Gray, Scott and Chess, Benjamin and Clark, Jack and Berner, Christopher and McCandlish, Sam and Radford, Alec and Sutskever, Ilya and Amodei, Dario},
 booktitle = {Advances in Neural Information Processing Systems},
 editor = {H. Larochelle and M. Ranzato and R. Hadsell and M.F. Balcan and H. Lin},
 pages = {1877--1901},
 publisher = {Curran Associates, Inc.},
 title = {Language Models are Few-Shot Learners},
 url = {https://proceedings.neurips.cc/paper_files/paper/2020/file/1457c0d6bfcb4967418bfb8ac142f64a-Paper.pdf},
 volume = {33},
 year = {2020}
}

@article{jiang2025coderl+,
  title={CodeRL+: Improving Code Generation via Reinforcement with Execution Semantics Alignment},
  author={Jiang, Xue and Dong, Yihong and Liu, Mengyang and Deng, Hongyi and Wang, Tian and Tao, Yongding and Cao, Rongyu and Li, Binhua and Jin, Zhi and Jiao, Wenpin and others},
  journal={arXiv preprint arXiv:2510.18471},
  year={2025}
}

@article{feng2025retool,
  title={Retool: Reinforcement learning for strategic tool use in llms},
  author={Feng, Jiazhan and Huang, Shijue and Qu, Xingwei and Zhang, Ge and Qin, Yujia and Zhong, Baoquan and Jiang, Chengquan and Chi, Jinxin and Zhong, Wanjun},
  journal={arXiv preprint arXiv:2504.11536},
  year={2025}
}

@article{yang2024qwen2,
  title={Qwen2. 5-math technical report: Toward mathematical expert model via self-improvement},
  author={Yang, An and Zhang, Beichen and Hui, Binyuan and Gao, Bofei and Yu, Bowen and Li, Chengpeng and Liu, Dayiheng and Tu, Jianhong and Zhou, Jingren and Lin, Junyang and others},
  journal={arXiv preprint arXiv:2409.12122},
  year={2024}
}

\clearpage
\appendix

\section{Reasoning Gym Experiment Hyperparameters}
\label{appendix:rg_hyperparameters}

This appendix details the hyperparameters used for training and evaluation on the Reasoning Gym benchmark environments.

\subsection{Training Hyperparameters}

\begin{table}[H]
\centering
\caption{Common training hyperparameters across all Reasoning Gym environments.}
\label{tab:rg_common_hyperparams}
\begin{tabular}{ll}
\toprule
\textbf{Hyperparameter} & \textbf{Value} \\
\midrule
\multicolumn{2}{l}{\textit{Model}} \\
Base Model & Qwen2.5-3B-Instruct \\
\midrule
\multicolumn{2}{l}{\textit{Optimization}} \\
Algorithm & GRPO / RLOO \\
Training Steps & 500 \\
Learning Rate & $1 \times 10^{-6}$ \\
LR Warmup & None \\
Gradient Clipping & 1.0 \\
\midrule
\multicolumn{2}{l}{\textit{PPO/GRPO}} \\
Train Batch Size & 32 \\
Mini-batch Size & 16 \\
Micro-batch Size (per GPU) & 4 \\
PPO Epochs & 1 \\
Clip Ratio ($\epsilon$) & 0.2 \\
Entropy Coefficient & 0.001 \\
\midrule
\multicolumn{2}{l}{\textit{KL Regularization}} \\
KL Coefficient & 0.001 \\
KL Loss Type & low\_var\_kl \\
\midrule
\multicolumn{2}{l}{\textit{Sampling}} \\
Temperature & 1.0 \\
Top-$p$ & 1.0 \\
Top-$k$ & Disabled \\
\midrule
\multicolumn{2}{l}{\textit{Infrastructure}} \\
Number of GPUs & 4xA6000 \\
Tensor Parallel Size & 4 \\
Parallelization Strategy & FSDP \\
\midrule
\multicolumn{2}{l}{\textit{Reward}} \\
Primary Reward & Accuracy \\
Format Reward Scaling & 0.2 \\
Developer Prompt & DeepSeekZero \\
\bottomrule
\end{tabular}
\end{table}

\begin{table}[H]
\centering
\caption{Environment-specific hyperparameters for Reasoning Gym experiments.}
\label{tab:rg_env_hyperparams}
\begin{tabular}{lccccc}
\toprule
\textbf{Hyperparameter} & \textbf{ARC-1D} & \textbf{Puzzle 24} & \textbf{Manipulate Matrix} & \textbf{Spell Backward} & \textbf{Word Sorting} \\
\midrule
Max Prompt Length & 4,096 & 1,024 & 3,072 & 768 & 1,024 \\
Max Response Length & 2,048 & 512 & 512 & 256 & 512 \\
Samples per Prompt ($n$) & 4 & 4 & 4 & 8 & 4 \\
\bottomrule
\end{tabular}
\end{table}

\begin{table}[H]
\centering
\caption{Training data consumption for Reasoning Gym experiments. Reasoning Gym environments generate problems procedurally, so we report the effective number of unique prompts seen during training.}
\label{tab:rg_data_consumption}
\begin{tabular}{lccccc}
\toprule
\textbf{Metric} & \textbf{ARC-1D} & \textbf{Puzzle 24} & \textbf{Manipulate Matrix} & \textbf{Spell Backward} & \textbf{Word Sorting} \\
\midrule
Training Steps & 500 & 500 & 500 & 500 & 500 \\
Batch Size & 32 & 32 & 32 & 32 & 32 \\
Samples per Prompt ($n$) & 4 & 4 & 4 & 8 & 4 \\
\bottomrule
\end{tabular}
\end{table}

\begin{table}[H]
\centering
\caption{Dataset generation configurations for each Reasoning Gym environment.}
\label{tab:rg_dataset_configs}
\begin{tabular}{ll}
\toprule
\textbf{Environment} & \textbf{Configuration Parameters} \\
\midrule
ARC-1D & \texttt{min\_size=10}, \texttt{max\_size=30}, \texttt{num\_train=3} \\
\midrule
Puzzle 24 & \texttt{operators=\{+, -, *, /\}}, \texttt{min\_value=1}, \texttt{max\_value=10} \\
\midrule
Manipulate Matrix & \texttt{min\_rows=2}, \texttt{max\_rows=10}, \texttt{min\_cols=2}, \texttt{max\_cols=10}, \\
& \texttt{min\_transforms=1}, \texttt{max\_transforms=10} \\
\midrule
Spell Backward & \texttt{min\_word\_len=3}, \texttt{max\_word\_len=10} \\
\midrule
Word Sorting & \texttt{min\_words=3}, \texttt{max\_words=10}, \\
& \texttt{min\_word\_length=3}, \texttt{max\_word\_length=12} \\
\bottomrule
\end{tabular}
\end{table}

\subsection{CBRL In-Context Learning Injection}

For experiments using Context Bootstrapped Reinforcement Learning (CBRL), we inject in-context learning examples during training with the following configuration:

\begin{table}[H]
\centering
\caption{CBRL in-context learning injection hyperparameters.}
\label{tab:rg_cbrl_hyperparams}
\begin{tabular}{ll}
\toprule
\textbf{Hyperparameter} & \textbf{Value} \\
\midrule
ICL Injection Enabled & True \\
Start Probability ($p_{\text{start}}$) & 0.5 \\
End Probability ($p_{\text{end}}$) & 0.0 \\
Number of ICL Examples & 2 \\
Injection Seed & 0 \\
\bottomrule
\end{tabular}
\end{table}

The injection probability is linearly annealed from $p_{\text{start}}$ to $p_{\text{end}}$ over the course of training, allowing the model to gradually transition from learning with demonstrations to independent problem-solving.

\subsection{Evaluation Hyperparameters}

\begin{table}[H]
\centering
\caption{Evaluation hyperparameters for Reasoning Gym experiments.}
\label{tab:rg_eval_hyperparams}
\begin{tabular}{ll}
\toprule
\textbf{Hyperparameter} & \textbf{Value} \\
\midrule
Temperature & 0.6 \\
Top-$p$ & 0.9 \\
Evaluation Set Size & 100 \\
Evaluation Repeats & 3 \\
Developer Prompt & DeepSeekZero \\
Precision & bfloat16 \\
\bottomrule
\end{tabular}
\end{table}
\newpage
\section{Q Programming Experiment Hyperparameters}
\label{appendix:qprog_hyperparameters}

This appendix details the hyperparameters used for training and evaluation on the Q programming code generation task.

\subsection{Training Hyperparameters}

\begin{table}[H]
\centering
\caption{Training hyperparameters for Q programming experiments.}
\label{tab:qprog_hyperparams}
\begin{tabular}{ll}
\toprule
\textbf{Hyperparameter} & \textbf{Value} \\
\midrule
\multicolumn{2}{l}{\textit{Model}} \\
Base Model & qqwen-7B-Pretrain~\cite{DBLP:journals/corr/abs-2508-06813} \\
Model Size & 7B parameters \\
\midrule
\multicolumn{2}{l}{\textit{Optimization}} \\
Algorithm & GRPO \\
Total Epochs & 64 \\
Learning Rate & $1 \times 10^{-6}$ \\
LR Warmup & None \\
Gradient Clipping & 1.0 \\
\midrule
\multicolumn{2}{l}{\textit{PPO/GRPO}} \\
Train Batch Size & 64 \\
Mini-batch Size & 16 \\
Micro-batch Size (per GPU) & 8 \\
PPO Epochs & 1 \\
Clip Ratio ($\epsilon$) & 0.2 \\
Entropy Coefficient & 0.0 \\
\midrule
\multicolumn{2}{l}{\textit{KL Regularization}} \\
KL Coefficient & 0.001 \\
KL Loss Type & low\_var\_kl \\
\midrule
\multicolumn{2}{l}{\textit{Sampling}} \\
Temperature & 1.0 \\
Top-$p$ & 1.0 \\
Samples per Prompt ($n$) & 8 \\
\midrule
\multicolumn{2}{l}{\textit{Sequence Lengths}} \\
Max Prompt Length & 4,096 \\
Max Response Length & 2,048 \\
\midrule
\multicolumn{2}{l}{\textit{Infrastructure}} \\
Number of GPUs & 4$\times$GH200 \\
Tensor Parallel Size & 4 \\
Parallelization Strategy & FSDP \\
\bottomrule
\end{tabular}
\end{table}

\subsection{Dataset and Training Data Consumption}

The Q programming task uses the \texttt{morganstanley/sft-python-q-problems} dataset, which contains Python programming problems paired with Q language solutions.

\begin{table}[H]
\centering
\caption{Dataset statistics and training data consumption for Q programming experiments.}
\label{tab:qprog_data}
\begin{tabular}{ll}
\toprule
\textbf{Metric} & \textbf{Value} \\
\midrule
\multicolumn{2}{l}{\textit{Dataset}} \\
Training Set Size & 542 problems \\
Test Set Size & 136 problems \\
Data Source & morganstanley/sft-python-q-problems \\
\midrule
\multicolumn{2}{l}{\textit{Training Data Consumption}} \\
Batch Size & 64 \\
Total Epochs & 64 \\
Total Training Steps & 512 \\
Samples per Prompt ($n$) & 8 \\
\bottomrule
\end{tabular}
\end{table}

\subsection{CBRL In-Context Learning Injection}

For experiments using Context Bootstrapped Reinforcement Learning (CBRL), we inject in-context learning examples during training:

\begin{table}[H]
\centering
\caption{CBRL in-context learning injection hyperparameters for Q programming.}
\label{tab:qprog_cbrl}
\begin{tabular}{ll}
\toprule
\textbf{Hyperparameter} & \textbf{Value} \\
\midrule
ICL Injection Enabled & True \\
Start Probability ($p_{\text{start}}$) & 0.5 \\
End Probability ($p_{\text{end}}$) & 0.0 \\
Number of ICL Examples & 2 \\
Injection Seed & 42 \\
\bottomrule
\end{tabular}
\end{table}

The injection probability is linearly annealed from $p_{\text{start}}$ to $p_{\text{end}}$ over the course of training.

\subsection{Reward Function}

The Q programming task uses a custom reward function that executes the generated Q code against test cases:

\begin{table}[H]
\centering
\caption{Reward function configuration for Q programming experiments.}
\label{tab:qprog_reward}
\begin{tabular}{ll}
\toprule
\textbf{Parameter} & \textbf{Value} \\
\midrule
Reward Type & Code Execution \\
Base Reward Weight & 1.0 \\
Perfect Bonus & 2.0 \\
Max Test Cases per Problem & 5 \\
Execution Timeout & 10 seconds \\
\midrule
\multicolumn{2}{l}{\textit{Reward Computation}} \\
Base Reward & $\frac{\text{passed tests}}{\text{total tests}} \times \text{base\_weight}$ \\
Perfect Bonus & +2.0 if all tests pass \\
\bottomrule
\end{tabular}
\end{table}

The reward function extracts Q code from the model output, executes it against up to 5 test cases, and computes a reward based on the pass rate with an additional bonus for perfect solutions.

\section{Training Prompts}
\label{appendix:prompts}

This appendix presents the exact prompts used during training for both the Reasoning Gym and Q Programming experiments.

\subsection{Reasoning Gym Prompt}

For Reasoning Gym experiments, we use the DeepSeekZero system prompt, which instructs the model to use a structured thinking and answer format.

\begin{tcolorbox}[
    colback=gray!5!white,
    colframe=gray!75!black,
    title=DeepSeekZero System Prompt,
    fonttitle=\bfseries
]
\begin{verbatim}
A conversation between User and Assistant. The user asks a
question, and the Assistant solves it. The assistant first
thinks about the reasoning process in the mind and then
provides the user with the answer. The reasoning process
and answer are enclosed within <think> </think> and
<answer> </answer> tags, respectively, i.e.,

<think> reasoning process here </think>
<answer>answer here</answer>

Do not explain your reasoning inside the answer tags,
provide only the final answer. When an example is provided,
you should strictly follow the format of the output/answer
in that example.
\end{verbatim}
\end{tcolorbox}

The user message contains the problem description specific to each Reasoning Gym environment (e.g., ARC-1D patterns, Puzzle 24 numbers, matrix transformations, etc.).

\subsubsection{CBRL Few-Shot Injection Format}

When CBRL in-context learning injection is enabled, few-shot examples are prepended to the conversation in the following format:

\begin{tcolorbox}[
    colback=blue!5!white,
    colframe=blue!75!black,
    title=Few-Shot Example Format (Reasoning Gym),
    fonttitle=\bfseries
]
\begin{verbatim}
[System] <DeepSeekZero system prompt>

[User] <Example problem 1>

[Assistant] <think><reasoning for example 1></think>
<answer><solution for example 1></answer>

[User] <Example problem 2>

[Assistant] <think><reasoning for example 2></think>
<answer><solution for example 2></answer>

[User] <Actual problem to solve>
\end{verbatim}
\end{tcolorbox}

\subsection{Q Programming Prompt}

For Q Programming experiments, we use a raw prompt format (without chat template) that instructs the model to generate Q code.

\begin{tcolorbox}[
    colback=gray!5!white,
    colframe=gray!75!black,
    title=Q Programming Prompt Template,
    fonttitle=\bfseries
]
\begin{verbatim}
Write a Q solve function that solves the following problem:

<problem description>

You must implement a function called 'solve' that accepts
the arguments specified in the problem description. The
function should return the correct output for the given
inputs.

Output ONLY the Q solve function inside a single fenced
code block. Do not include any other text, explanations,
or test harnesses.

Example format:
```q
solve:{[args] // your implementation here}
```
\end{verbatim}
\end{tcolorbox}

\subsubsection{CBRL Few-Shot Injection Format}

When CBRL in-context learning injection is enabled for Q Programming with raw prompt mode, the few-shot example problem descriptions are concatenated with the target problem:

\begin{tcolorbox}[
    colback=blue!5!white,
    colframe=blue!75!black,
    title=Few-Shot Example Format (Q Programming),
    fonttitle=\bfseries
]
\begin{verbatim}
Write a Q solve function that solves the following problem:

<Example 1>

<Example 2>

<Actual problem to solve>

You must implement a function called 'solve' that accepts
the arguments specified in the problem description. The
function should return the correct output for the given
inputs.

Output ONLY the Q solve function inside a single fenced
code block. Do not include any other text, explanations,
or test harnesses.

Example format:
```q
solve:{[args] // your implementation here}
```
\end{verbatim}
\end{tcolorbox}

\subsection{Expected Output Format}

\begin{table}[H]
\centering
\caption{Expected model output formats for each task.}
\label{tab:output_formats}
\begin{tabular}{lp{10cm}}
\toprule
\textbf{Task} & \textbf{Expected Output Format} \\
\midrule
Reasoning Gym & \texttt{<think>}reasoning process\texttt{</think>}\newline\texttt{<answer>}final answer\texttt{</answer>} \\
\midrule
Q Programming & \texttt{\textasciigrave\textasciigrave\textasciigrave q \newline
solve:{[args]}\textasciigrave\textasciigrave\textasciigrave} \\
\bottomrule
\end{tabular}
\end{table}

\end{document}